\documentclass[journal]{IEEEtran}
\usepackage{graphicx}
\usepackage{float}
\usepackage{caption}
\usepackage{amsmath}
\pdfoutput=1
\usepackage[linesnumbered, ruled]{algorithm2e}
\usepackage{amsfonts}
\usepackage{amssymb}
\usepackage{bm}
\usepackage{cite}
\usepackage{color}
\usepackage{multirow}
\usepackage{tabularx}
\newtheorem{lemma}{Lemma}
\newtheorem{theorem}{\bf Theorem}
\newtheorem{prob}{Problem}
\newtheorem{assumption}{\it Assumption}
\newtheorem{rem}{\bf Remark}
\usepackage{times, epsfig}
\usepackage{subfig}

\usepackage{booktabs}% To describe the three line table

\everymath{\displaystyle}
\def\BibTeX{{\rm B\kern-.05em{\sc i\kern-.025em b}\kern-.08em
    T\kern-.1667em\lower.7ex\hbox{E}\kern-.125emX}}

\begin{document}

\title{ Federated Learning Robust to Byzantine Attacks: Achieving Zero Optimality Gap}

\author{
Shiyuan Zuo, Rongfei Fan, Han Hu, Ning Zhang, and Shimin Gong
%Author 1, Author 2, Author 3, Author 4, and Author 5
 \vspace{-5 mm}
\thanks{%Author 1, Author 2, and Author 3 are with Affiliation A;
%Author 4 is with Affiliation B;
%Author 5 is with Affiliation C.
S. Zuo and R. Fan are with the School of Cyberspace Science and Technology, Beijing Institute of Technology, Beijing 100081, P.R. China.
%(e-mail:  \{3120210836, fanrongfei\}@bit.edu.cn). 
H. Hu is with the School of Information and Electronics, Beijing Institute of Technology, Beijing 100081, P. R. China.
%(e-mail: hhu@bit.edu.cn). 
N. Zhang is with the Department of Electrical and Computer Engineering, University of Windsor, Windsor, ON N9B 3P4, Canada.
%(e-mail: ning.zhang@uwindsor.ca). 
S. Gong is with the School of Intelligent Systems Engineering, Sun Yat-sen University, Shenzhen 518055,  China.
%(email: gongshm5@mail.sysu.edu.cn) 
}
}

\maketitle

\begin{abstract}
In this paper, we propose a robust aggregation method for federated learning (FL) that can effectively tackle malicious Byzantine attacks. At each user, model parameter is firstly updated by multiple steps, which is adjustable over iterations, and then pushed to the aggregation center directly. This decreases the number of interactions between the aggregation center and users, allows each user to set training parameter in a flexible way, and reduces computation burden compared with existing works that need to combine multiple historical model parameters. At the aggregation center, geometric median is leveraged to combine the received model parameters from each user. Rigorous proof shows that zero optimality gap is achieved by our proposed method with linear convergence, as long as the fraction of Byzantine attackers is below half. Numerical results verify the effectiveness of our proposed method.
%'s effectiveness and the advantage over benchmark methods.
%the effectiveness of our proposed method and 
%Numerical results demonstrate the benefits of flexible setup for local update steps and the communication efficiency of our proposed method. Moreover, it can achieve a higher training performance in terms of training loss, test accuracy, and convergence speed, over benchmark methods.
\end{abstract}


%\vspace{-5 mm}

\section{Introduction} \label{sec:intro}
Federated learning (FL) is a distributed machine learning framework aiming at addressing data privacy issue \cite{konevcny2016federated, marano2013nearest, yu2017distributed}, which is realized through iterative local training and aggregation of model parameters or local gradients between distributed users and the aggregation center until convergence.
%In FL, distributed users first train their own data samples privately and then push the model parameters or most up-to-date local gradients to the aggregation center, which will perform aggregation and feed back the aggregated one to eiteratively until convergence.
%for aggregation. 
%In FL, distributed users train local models based on their own data samples without sharing them with others. They then push something about the model parameters, such as the model parameters or most up-to-date local gradients, to an aggregation center for parameter aggregation. The aggregation center then broadcasts the aggregated result to the users to continue the local training process. This iterative interaction between distributed users and the aggregation center continues over multiple rounds until convergence is achieved.
Although being privacy protective, FL is still vulnerable to random attacks.
% and data errors from users. 
For instance, malicious users may attempt to disrupt the FL process by sending corrupted or misleading messages to the aggregation center. This type of attacks actually falls under the category of Byzantine attacks.
%, which could severely interfere with the FL process.
It is vital to overcome Byzantine attacks, but it is also challenging to do so as Byzantine attackers could be adaptive and inject arbitrary messages to severely interfere the FL process
%by compromising devices in the FL system
\cite{lamport2019byzantine}.

In the literature, several methods have been proposed to address the vulnerability of FL against Byzantine attacks \cite{xie2018generalized,yin2018byzantine,su2019securing,pillutla2022robust,wu2020federated,turan2022robust}.
In \cite{xie2018generalized,yin2018byzantine,su2019securing}, median, trimmed median, and iterative filtering are utilized at the aggregation center to aggregate uploaded stochastic gradient descent (SGD), so as to tolerate a certain number (at most 15\% in numerical results) of Byzantine attackers.
To be robust to more Byzantine attackers, geometric median of each user's updated vector are utilized in \cite{pillutla2022robust,wu2020federated,turan2022robust}.
%The difference among \cite{pillutla2022robust,wu2020federated,turan2022robust} lies in the vector uploaded by each user.
Specifically, the Robust Federated Aggregation (RFA) algorithm in \cite{pillutla2022robust} allows each user to perform multiple steps of local updating with uniform setup, and takes the average of the model parameters generated in these steps as the updating vector.
In contrast, \cite{wu2020federated} and \cite{turan2022robust} only run one step of local updating for each user.
The Stochastic Average Gradient Algorithm (SAGA) in \cite{wu2020federated} takes the summation of local gradient gap for succesive two iterations and the average of all the local gradients in historical iterations as the update vector.
The Robust Aggregating Normalized Gradient Method (RANGE) in \cite{turan2022robust}
%The authors in \cite{turan2022robust} proposed a Robust Aggregating Normalized Gradient Method (RANGE), which 
takes the geometric median of local gradients in multiple historical iterations as the updating vector.

For \cite{pillutla2022robust,wu2020federated,turan2022robust}, their proposed methods have been shown to converge so long as the proportion of Byzantine attackers is less than half, but can hardly achieve zero optimality gap.
%However, zero optimality gap can be hardly achieved.
Moreover, \cite{wu2020federated,turan2022robust} do not assume multiple steps of local updating, necessitating frequent interaction with the aggregation center, which would incur heavy communication overhead.
In contrast, \cite{pillutla2022robust} allows for multiple steps of local updating but with a uniform setup, which may not be generic to all use cases and thus lead to the loss of training performance.
Last but not least, the preprocessing of local/global gradients or model parameters at each user before pushing a vector to the aggregation center involves algorithmic mean or even geometric mean, which can be computationally inefficient.

To overcome the above drawbacks, we propose a novel aggregation method for FL that can effectively handle the Byzantine attacks while achieving zero optimality gap.
Our contributions are given as follows:
\begin{itemize}
\item Firstly, we propose a flexible and efficient aggregation method that allows each user to perform multiple local updates with a variable step size per iteration, thus enabling efficient communication between the users and the aggregator. Unlike previous methods, our approach requires no preprocessing of model parameters or local gradients before uploading, contributing to significant computational savings.
\item Secondly, we prove that our method achieves zero optimality gap at a linear convergence rate, and is highly robust to Byzantine attacks with strong guarantees even when up to half of the users are attackers. These results represent a significant improvement over existing methods, which can hardly  achieve zero optimality gap.
\item Finally, through numerical experiments, we demonstrate the superior performance of our method %showing that it outperforms
over existing benchmark methods in terms of training loss, test accuracy, and convergence speed across a variety of settings.
\end{itemize}

\section{System Model} \label{sec:model}

Consider a FL system with $M$ users, denoted as $\mathcal{M} \triangleq \{1, 2, ..., M\}$, and one aggregation center.
%These $M$ users constitute the set $\mathcal{M} \triangleq \{1, 2, ..., M\}$.
At the $m$th user, it has a dataset $\mathcal{S}_m$, consisting of $S_m$ ground-true labels.
For the $s$th ground-true label in $\mathcal{S}_m$,  it is composed of a pair of data point $\bm{x}_{m,s}$ and $\bm{y}_{m,s}$, which represent the input vector and output vector, respectively.
The objective of the FL system is to learn a model parameter vector $\bm{w} \in \mathcal{R}^{p}$ so as to minimize the loss function for approximating the data labels of all the users, denoted as $F(\bm{w})$, through cooperation between each user and the aggregation center. In other words, we need to solve the following problem

\begin{prob} \label{p:min_F_w}
\begin{small}
\begin{equation*}
\mathop{\min}_{\bm{w}} F(\bm{w})
\end{equation*}
\end{small}
\end{prob}
The loss function $F(\bm{w})$ in Problem \ref{p:min_F_w} is defined as
%\begin{small}
%\begin{equation}
%  \begin{aligned}
$F(\bm{w})
     \triangleq \frac{1}{\sum_{m=1}^M  S_m }  \sum_{m=1}^M  \sum_{s \in \mathcal{S}_m}  f(\bm{w}, \bm{x}_{m,s}, \bm{y}_{m,s})$, 
%\end{aligned}
%\end{equation}
%\end{small}
where $f(\bm{w}, \bm{x}_{m,s}, \bm{y}_{m,s})$ is the loss function for approximating the label $\left(\bm{x}_{m,s}, \bm{y}_{m,s}\right)$ with given parameter vector $\bm{w}$.
For the ease of discussion, we define the local loss function for the $m$th user as
%\begin{small}
%\begin{equation}
$F_m(\bm{w}) \triangleq \frac{1}{S_m} \sum_{s \in \mathcal{S}_m} f(\bm{w}, \bm{x}_{m,s}, \bm{y}_{m,s})$,
%\end{equation}
%\end{small}
and then the loss function $F(\bm{w})$ can be also written as
%\begin{small}
%\begin{equation}
$F(\bm{w}) = \sum_{m \in \mathcal{M}} \frac{S_m}{\sum_{m=1}^M S_m} F_m(\bm{w})$.
%\end{equation}
%\end{small}

%Among these $U$ users, there are $B$ users belonging to Byzantine attackers.
% \subsection{FL Optimization under Byzantine Attacks} \label{subsec:FL optimi}

% \subsection{Assumptions}
For the loss function $F(\bm{w})$ and $F_m(\bm{w})$, the following assumptions are made, which are common in related works {\cite{pillutla2022robust}}.
\begin{assumption}[Smoothness] \label{ass:smooth}
  The loss function $F(\bm{w})$ has L-Lipschitz continuous gradients, i.e., for $\forall \bm{w}_1, \bm{w}_2 \in \mathcal{R}^p$, there is
  \begin{small}
  \begin{equation}
    \begin{aligned}
      F(\bm{w}_1) - F(\bm{w}_2) \leqslant \nabla F(\bm{w}_2)^T(\bm{w}_1 - \bm{w}_2) + \frac{L}{2}\left\lVert \bm{w}_1 - \bm{w}_2 \right\rVert^2
    \end{aligned}
  \end{equation}
  \end{small}
where $\nabla F(\cdot)$ represents the gradient vector of function $F(\cdot)$.
\end{assumption}

\begin{assumption}[Strong Convexity]\label{ass:convex}
  The loss function $F(\bm{w})$ is $\mu$-strong convex, i.e., for $\forall \bm{w}_1, \bm{w}_2 \in \mathcal{R}^p$, there is
  \begin{small}
  \begin{equation}
    (\nabla F(\bm{w}_1) - \nabla F(\bm{w}_2))^T (\bm{w}_1 - \bm{w}_2) \geqslant \mu \left\lVert \bm{w}_1 - \bm{w}_2 \right\rVert ^2
  \end{equation}
  \end{small}
\end{assumption}

\begin{assumption}[Unbiased Gradient] \label{ass:unbias}
For any $m\in \mathcal{M}$,
suppose a subset of dataset $\mathcal{S}_m$ is selected randomly, which is denoted as $\xi_m$, %\textcolor{red}{with training data $\xi_m$},
define
\begin{small}
\begin{equation}
F_m(\bm{w}; \xi_m) \triangleq \frac{1}{| \xi_m |} \sum_{s \in \xi_m} f(\bm{w}, \bm{x}_{m,s}, \bm{y}_{m,s}),
\end{equation}
\end{small}
the unbiased gradient implies that
%denote $F(\bm{w}; \xi_m) $, then there is
%for any fixed model parameter $\bm{w}$, the stochastic gradient \textcolor{red}{$\nabla F(\bm{w}_{m})$} is an unbiased estimator of the true gradient corresponding to the parameter $\bm{w}$, namely,
\begin{small}
\begin{equation}
  \mathbb{E} \left \{ \nabla F_{m}(\bm{w};\xi_m) \right \} = \nabla F(\bm{w}), \forall m \in \mathcal{M}.
\end{equation}
\end{small}
% where $\bm{\xi} = \{\xi_1, \xi_2, ..., \xi_M \}$ and
% \begin{equation}
% F(\bm{w}, \bm{\xi}) \triangleq \sum_{m \in \mathcal{M}} \frac{|\xi_m|}{\sum_{m=1}^M |\xi_m|} F_m(\bm{w},\xi_m).
% \end{equation}
\end{assumption}

\begin{assumption}[Bounded Variance] \label{ass:bound}
  The variance of $\nabla F_m(\bm{w};\xi_m)$ with respect to the norm of its expectation is upper-bounded by $\delta$, i.e.,
  \begin{small}
  \begin{equation}
    \frac{\mathbb{E} \left \{ \left\lVert \nabla F_m(\bm{w};\xi_m) - \nabla F(\bm{w}) \right\rVert^2 \right \}}{\left\lVert \nabla F(\bm{w}) \right\rVert^2} \leqslant \delta^2, \forall m \in \mathcal{M}, \bm{w} \in \mathcal{R}^p
  \end{equation}
  \end{small}
\end{assumption}

%In a traditional FL, with Assumption 1, 2, ,3, optimal solution of Problem \ref{p:min_F_w} can be achieved through iterative interaction between the group of $M$ users and the aggregation center.
%However, among these $M$ users, a subset of them

In a traditional FL, iterative interactions between the group of $M$ users and the aggregation center are performed to update the model parameter $\bm{w}$ until convergence.
For the $t$th iteration, it is performed as follows:
%\begin{itemize}
% \item 
\textbf{Step 1 (Local Update)}
Every user, say the $m$th user, calculates its updated parameter vector, denoted as $\bm{w}_m^t$, based on its local dataset $\mathcal{S}_m$ and the the global parameter vector broadcasted by the aggregation center in the last iteration, denoted as $\bm{w}^{t-1}$. The updating from $\bm{w}^{t-1}$ to $\bm{w}_m^t$ is usually made through gradient method {\cite{konevcny2016federated}};
%\item 
\textbf{Step 2 (Aggregation and Broadcasting)}
Each user sends its most updated parameter vector $\bm{w}_{m}^t$ to the aggregation center. Then the aggregation center combines all the received local parameter vectors into a common vector, known as global parameter vector denoted as $\bm{w}^{t}$, and then broadcasts it to every user.
%\end{itemize}

With Assumptions \ref{ass:smooth}, \ref{ass:convex}, \ref{ass:unbias}, and \ref{ass:bound}, the above iterative operation can help to reach the optimal solution of Problem \ref{p:min_F_w} if all users provide truthful local model parameters to the aggregation center in each iteration according to {\cite{8737464}}.
In real application, however, some users may be corrupted and intentionally disrupt the aggregation process. These malicious users, known as Byzantine users, can upload fake and arbitrarily large vectors as a replacement for their actual local model parameter vectors to the aggregation center, leading to a failure in converging to the optimal solution of Problem \ref{p:min_F_w}.
In contrast, honest users refer to uncorrupted users and are used for ease of discussion in the subsequent analysis.

%Denote the set of Byzantine users as $\mathcal{B}$, which includes $B$ users.
%Suppose $B$ is less than half of the crew members in the FL system.
%In other words, there is $B < {M}/{2}$.
%This assumption is general and can be usually seen in literature {\cite{pillutla2022robust, turan2022robust, wu2020federated}}.
%Suppose $\bm{z}_{m}^t$ is the real vector uploaded by the $m$th user to the aggregation center in $t$th iteration, then for $m \in \mathcal{B}$, there is $\bm{z}_m^t = \star$, where $\bm{\star }\in \mathcal{R} ^p$ denotes an arbitrary vector.
Denote the set of Byzantine users as $\mathcal{B}$, which includes $B$ users.
Suppose $B < {M}/{2}$, which is general and can be usually seen in literature {\cite{pillutla2022robust, turan2022robust, wu2020federated}}.
Suppose $\bm{z}_{m}^t$ is the real vector uploaded by the $m$th user to the aggregation center in $t$th iteration, then for $m \in \mathcal{B}$, there is $\bm{z}_m^t = \star$, where $\bm{\star }\in \mathcal{R} ^p$ denotes an arbitrary vector.

\section{Proposed Aggregation Method and Convergence Analysis}

\subsection{Proposed Aggregation Method} \label{subsec:distriSGD}

%To reduce communication burden due to frequent interactions between $M$ users and the aggregation center and overcome the attacks of Byzantine users, 
We propose to run multiple updates of local model parameter and aggregate all the uploaded vectors in a robust way.
Specifically, in $t$th iteration
\begin{itemize}
\item
In Step 1 (Local Update), local model parameter is updated $K^t$ times before it is uploaded to the aggregation center for honest users $m \in \mathcal{M} \setminus \mathcal{B}$.
Denote $\bm{w}_m^{t, k}$ as the local model parameter of the $m$th user for the $k$th update and $\eta_m^{t,k}$ as the associated learning rate, then by following SGD method {\cite{gower2019sgd}}, there is
%\begin{small}
%\begin{equation} \label{equ:update}
  $\bm{w}_m^{t, k} = \bm{w}_m^{t, k - 1} - \eta_m^{t, k} \nabla F_m(\bm{w}_m^{t, k - 1}; \xi_m^{t, k}), k=1, ..., K^t$
%\end{equation}
%\end{small}
where
$\xi_m^{t, k}$ is a randomly selected subset of the data sample set $\mathcal{S}_m$ and $\bm{w}_m^{t, 0} = \bm{w}^t$.
With the above definition,
$\bm{w}_m^{t, K^t}$ can be expressed as
%\begin{small}
%\begin{equation}
 $ \bm{w}_m^{t, K^t} = \bm{w}^t - \sum_{k = 1}^{K^t} \eta_m^{t, k}\nabla F_m(\bm{w}_m^{t, k - 1}; \xi_m^{t, k})$.
%\end{equation}
%\end{small}
Set $\bm{z}_m^t = \bm{w}_m^{t, K^t}$ for $\forall m \in \mathcal{M} \setminus \mathcal{B}$.
\item
In Step 2 (Aggregation and Broadcasting),
the $m$th user uploads vector $\bm{z}_m^t$ to the aggregation center for $m \in \mathcal{M}$.
To mitigate the effects of Byzantine attacks,
we take the geometric median of all the uploaded vectors, which is given as
%\begin{small}
%\begin{align} \label{e:geometric_mean}
$\bm{w}^{t + 1}
    = {\rm geomed}\{\bm{z}_1^t, \bm{z}_2^t,\dots, \bm{z}_M^t\}  \triangleq \arg \min_{\bm{z}^t} \sum_{ m \in \mathcal{M}} \left\lVert \bm{z}^t - \bm{z}_m^t  \right\rVert_2 $.
%\end{align}
%\end{small}
To work out the geometric median given above, which is essentially a convex optimization problem, Weiszfeld algorithm can be resorted to {\cite{aftab2014generalized}}.
After taking the geometric median, the aggregation center broadcasts $\bm{w}^{t +1}$ to all the users, in preparation for gradient calculation in $(t+1)$th iteration.
\end{itemize}

\subsection{Convergence Analysis}\label{sec:conver}

%In this subsection, the convergence of our proposed aggregation method is proved.
We first consider a simple case, when $K^t= K$ and $\eta_m^{t,k}= \eta$ for every $t$, $m$, and $k$,%The discussion for this case could pave the way for the discussion in a more general case.
%Also t
% This case 
which is in coordination with the setup in {\cite{pillutla2022robust, zhou2018convergence}}.
Associated convergence result is given as follows

\begin{theorem} \label{theo:fix}
With $K^t$ set as $K$ and $\eta_m^{t,k}$ set as $\eta$ for every $t$, $m$, and $k$, the optimality gap $\mathbb{E}  \{ F(\bm{w}^{t + 1}) - F(\bm{w}^*) \}$ is bounded in (\ref{equ:theo1}) as follows
\begin{small}
  \begin{align}
    \mathbb{E} \left \{ F(\bm{w}^{t + 1}) - F(\bm{w}^*) \right \}
    &\leqslant \frac{L}{2} (\gamma^K)^t C_{\beta}^{2t} \left\lVert \bm{w}^1 - \bm{w}^* \right\rVert^2 \label{equ:theo1}
  \end{align}
  \end{small}
  where $\bm{w}^*$ denotes the optimal solution of Problem \ref{p:min_F_w}, $\gamma = 1 - 2\eta\mu + \eta^2L^2(1+\delta^2)$, $\beta = {B}/{M}$ and $C_{\beta} = (2-2\beta)/(1-2\beta)$.
\end{theorem}

\begin{IEEEproof}
  %Please refer to Appendix \ref{app:fix}.
  The proof of Theorem \ref{theo:fix} relies on the following two lemmas.

  \begin{lemma} \label{lem:geom}
    Let $\mathcal{Z} = \{\bm{z}_1, \bm{z}_2, ..., \bm{z}_{|\mathcal{Z}|}\}$ be a subset of random vectors distributed in a {normed vector space}. If $\mathcal{Z}^{'}\subseteq   \mathcal{Z} $ and $\left\lvert \mathcal{Z}^{'} \right\rvert < {\left\lvert \mathcal{Z} \right\rvert }/{2}$, there is
    \begin{small}
    \begin{align}\label{unequ:gemo}
      \mathbb{E} \left \{ {\left\lVert \mathop{{\rm geomed}} \limits \{ \mathcal{Z} \} \right\rVert^2} \right \}
 %    \quad \quad \quad \quad
     \leqslant  \frac{\left(C_{\alpha }\right)^2}{\left\lvert \mathcal{Z} \right\rvert - \left\lvert \mathcal{Z}^{'} \right\rvert} \sum_{\bm{z}'_i \in \mathcal{Z} \setminus \mathcal{Z}^{'}} \left\lVert \bm{z}'_i \right\rVert^2
    \end{align}
    \end{small}
    where $i = 1, 2, \dots, n$ , $\alpha = {\left\lvert \mathcal{Z}^{'} \right\rvert}/{\left\lvert \mathcal{Z} \right\rvert} $ and $C_{\alpha} = {(2 - 2\alpha)}/{(1 - 2\alpha)}$.
  \end{lemma}

  \begin{IEEEproof}
    Please refer to {Lemma 2.1} of \cite{minsker2015geometric}.
  \end{IEEEproof}

  \begin{lemma}\label{lem:bound}
With $\gamma = 1 - 2\eta\mu + \eta^2L^2(1 + \delta^2)$ and $\bm{w}^*$ being the optimal solution of Problem \ref{p:min_F_w}, for any honest user $m$, i.e., $\forall m \in \mathcal{M} \setminus \mathcal{B}$,
% and $\bm{w}^*$ represents the optimal solution of $F(\bm{w})$,
the term $\mathbb{E} \{ \left\lVert \bm{w}_{m}^{t, K} - \bm{w}^* \right\rVert^2 \}$ can be bounded as
\begin{small}
    \begin{equation}
      \mathbb{E} \left \{ \left\lVert \bm{w}_{m}^{t, K} - \bm{w}^* \right\rVert^2 \right \} \leqslant (\gamma)^K \mathbb{E} \left \{ \left\lVert \bm{w}^t - \bm{w}^* \right\rVert^2 \right \}
    \end{equation}
    \end{small}
  \end{lemma}

  \begin{IEEEproof}
 % Please refer to Appendix \ref{app:bound}.
%Recalling (\ref{equ:update}) and
With Assumptions \ref{ass:smooth} and \ref{ass:convex}, there is
\begin{small}
  \begin{subequations} \label{e:Theo1_large_ineq}
    \begin{align}
      &\quad \mathbb{E} \left \{ \left\lVert \bm{w}_m^{t, K} - \bm{w}^* \right\rVert^2 \right \} \nonumber \\
      &= \mathbb{E} \left \{ \left\lVert \bm{w}_m^{t, K-1} - \eta \nabla F_m(\bm{w}_m^{t, K-1};\xi_m^{t, K-1}) - \bm{w}^* \right\rVert^2 \right \} \\
      &= \mathbb{E} \left \{ \left\lVert \bm{w}_m^{t, K-1} - \bm{w}^* \right\rVert^2 \right \}  + \mathbb{E} \left \{ \left\lVert \eta \nabla F_m(\bm{w}_m^{t, K-1};\xi_m^{t, K-1}) \right\rVert^2 \right \} \nonumber \\
      &\quad - 2\mathbb{E} \left \{ \left( \bm{w}_m^{t, K-1} - \bm{w}^* \right)^T \left( \eta \nabla F_m(\bm{w}_m^{t, K-1};\xi_m^{t, K-1}) \right) \right \} \\
      &= \mathbb{E} \left \{ \left\lVert \bm{w}_m^{t, K-1} - \bm{w}^* \right\rVert^2 \right \} + \mathbb{E} \left \{ \left\lVert \eta \nabla F_m(\bm{w}_m^{t, K-1};\xi_m^{t, K-1}) \right\rVert^2 \right \} \nonumber \\
      &\quad - \mathbb{E} \left \{ \eta \left( \bm{w}_m^{t, K-1} - \bm{w}^* \right)^T  \cdot \left( \nabla F(\bm{w}_m^{t, K-1}) - \nabla F(\bm{w}^*) \right) \right \}  \label{equ:ustrong} \\
      &\leqslant \mathbb{E} \left \{ \left\lVert \bm{w}_m^{t, K-1} - \bm{w}^* \right\rVert^2 \right \}  + \mathbb{E} \left \{ \left\lVert \eta \nabla F_m(\bm{w}_m^{t, K-1};\xi_m^{t, K-1}) \right\rVert^2 \right \} \nonumber \\
      &\quad -2 \mathbb{E} \left \{ \eta \mu\left\lVert \bm{w}_m^{t, K-1} - \bm{w}^* \right\rVert^2 \right \} \label{inequ:ustrong} \\
      &= \left( 1-2\eta \mu \right) \mathbb{E} \left \{ \left\lVert \bm{w}_m^{t, K-1} - \bm{w}^* \right\rVert^2 \right \} + \eta^2 \mathbb{E} \left \{ \left\lVert \nabla F(\bm{w}_m^{t, K-1}) \right\rVert^2 \right \}  \nonumber \\
      &\quad + \eta^2 \mathbb{E} \left \{ \left\lVert \nabla F_m(\bm{w}_m^{t, K-1};\xi_m^{t, K-1}) - \nabla F(\bm{w}_m^{t, K-1}) \right\rVert^2 \right \}  \\
   %   &\quad + \eta^2 \mathbb{E} \left \{ \left\lVert \nabla F(\bm{w}_m^{t, K-1}) \right\rVert^2 \right \} \\
      &\leqslant \left( 1-2\eta \mu \right) \mathbb{E} \left \{ \left\lVert \bm{w}_m^{t, K-1} - \bm{w}^* \right\rVert^2 \right \} \nonumber \\
      &\quad + \eta^2 (1+\delta^2) \mathbb{E} \left \{ \left\lVert \nabla F(\bm{w}_m^{t, K-1}) \right\rVert^2 \right \} \label{inequ:ass4} \\
      &= \left( 1-2\eta \mu \right) \mathbb{E} \left \{ \left\lVert \bm{w}_m^{t, K-1} - \bm{w}^* \right\rVert^2 \right \} \nonumber \\
      &\quad + \eta^2 (1+\delta^2) \mathbb{E} \left \{ \left\lVert \nabla F(\bm{w}_m^{t, K-1}) - \nabla F(\bm{w}^*) \right\rVert^2 \right \} \label{equ:lsmooth}\\
      &\leqslant \left( 1 - 2\eta \mu + \eta^2 L^2(1+\delta^2) \right) \mathbb{E} \left \{ \left\lVert \bm{w}_m^{t, K-1} - \bm{w}^* \right\rVert^2 \right \} \label{inequ:lsommth}
    \end{align}
  \end{subequations}
  \end{small}
%  where
%  \begin{itemize}
%    \item the two equalities in (\ref{equ:ustrong}) and (\ref{equ:lsmooth}) hold due to $\nabla F(\bm{w}^*) = 0$ and Assumption \ref{ass:unbias};
%    \item the inequality in (\ref{inequ:ustrong}) comes from Assumption \ref{ass:convex} because $\left( \bm{w}_m^{t, K-1} - \bm{w}^* \right)^T \left( \nabla F(\bm{w}_m^{t, K-1}) - \nabla F(\bm{w}^*) \right) \geqslant \\ \mu \notag \cdot \left\lVert \bm{w}_m^{t, K-1} - \bm{w}^* \right\rVert^2$.
%    \item the inequality in (\ref{inequ:lsommth}) is established  by the fact $\left\lVert \nabla F(\bm{w}_m^{t, K-1}) - \nabla F(\bm{w}^*) \right\rVert^2 \leqslant L^2 \notag \cdot \left\lVert \bm{w}_m^{t, K-1} - \bm{w}^* \right\rVert^2$ with the support of Assumption \ref{ass:smooth}
%    \item the inequality in (\ref{inequ:ass4}) is derived from Assumption \ref{ass:bound}.
%  \end{itemize}

With  $\gamma = 1 - 2\eta\mu + \eta^2L^2(1+\delta^2)$ and recall (\ref{e:Theo1_large_ineq}) for $K$ steps of local updates, we have
\begin{small}
  \begin{equation}
    \mathbb{E} \left \{ \left\lVert \bm{w}_m^{t, K} - \bm{w}^* \right\rVert^2 \right \} \leqslant (\gamma)^K \mathbb{E} \left \{ \left\lVert \bm{w}^t - \bm{w}^* \right\rVert^2 \right \}
  \end{equation}
  \end{small}

  This completes the proof of Lemma \ref{lem:bound}.
  \end{IEEEproof}

  Owing to the Assumption \ref{ass:smooth}, there is
  \begin{small}
%  \begin{subequations} \label{e:theo1_F_w}
%    \begin{align}
\begin{equation}
\begin{array}{ll}  \label{e:theo1_F_w}
      F(\bm{w}^{t + 1}) - F(\bm{w}^*)
       \leqslant \nabla F(\bm{w}^*)^T(\bm{w}^{t + 1} - \bm{w}^*) + \frac{L}{2}\left\lVert \bm{w}^{t + 1} - \bm{w}^* \right\rVert^2 \\
      = \frac{L}{2}\left\lVert \bm{w}^{t + 1} - \bm{w}^* \right\rVert^2
      \end{array}
      \end{equation}
%  \end{align}
%  \end{subequations}
  \end{small}
  %which holds since $\nabla F(\bm{w}^*) = 0$.

 Due to the randomness of training data subset $\xi_m^{t, k}$, we investigate how $\mathbb{E} \{ \left\lVert \bm{w}^{t + 1} - \bm{w}^* \right\rVert^2 \}$ varies with $t$. According to Lemma \ref{lem:geom} and Lemma \ref{lem:bound}, there is
 %the inequality of $\mathbb{E} \left \{ \left\lVert \bm{w}^{t + 1} - \bm{w}^* \right\rVert^2 \right \}$ can be shown,
 \begin{small}
%  \begin{subequations}
%    \begin{align}
\begin{equation}
\begin{array}{ll}
       \mathbb{E} \left \{ \left\lVert \bm{w}^{t + 1} - \bm{w}^* \right\rVert^2 \right \}
      \leqslant \frac{{(C_{\beta})^2}}{M - B} \sum\limits_{m \in \mathcal{M} \setminus \mathcal{B}} \mathbb{E} \left \{ \left\lVert \bm{w}_m^{t, K} - \bm{w}^* \right\rVert^2  \right \} \\
      \leqslant {(C_{\beta})^2} (\gamma)^K \mathbb{E} \left \{ \left\lVert \bm{w}^t - \bm{w}^* \right\rVert^2  \right \} \label{inequ:wt1}
% \end{align}
% \end{subequations}
\end{array}
\end{equation}
 \end{small}
{where $\beta = {B}/{M} \in [0, 1/2)$ since $B < M/2$,  and $C_{\beta} = \left({2-2\beta}\right)/\left({1-2\beta}\right) \geq 2$}.
 Thereafter, we have
% Then recalling the expression given in (\ref{inequ:wt1}), we have
\begin{small}
 \begin{align}
   \mathbb{E} \left \{ \left\lVert \bm{w}^{t + 1} - \bm{w}^* \right\rVert^2 \right \} \leqslant (\gamma)^{tK} (C_{\beta})^{2t} \mathbb{E} \left \{ \left\lVert \bm{w}^1 - \bm{w}^* \right\rVert^2 \right \} \label{inequ:Ew1}
  \end{align}
  \end{small}
Combining (\ref{e:theo1_F_w}) and (\ref{inequ:Ew1}), we have
\begin{small}
  \begin{align}
    \mathbb{E} \left \{ F(\bm{w}^{t + 1}) - F(\bm{w}^*) \right \}
    &\leqslant \frac{L}{2} (\gamma)^{tK} (C_{\beta})^{2t} \left\lVert \bm{w}^1 - \bm{w}^* \right\rVert^2
  \end{align}
  \end{small}

This completes the proof of Theorem \ref{theo:fix}.
%Here Theorem \ref{theo:fix} is proved.
\end{IEEEproof}

\begin{rem} \label{r:mark1}
%\begin{itemize}
 %\item 
\textbf{1)} Theorem \ref{theo:fix} substantiates that our proposed aggregation method can achieve zero optimality gap when $(\gamma)^{tK} (C_{\beta})^{2t}$ approaches zero as $t$ tends to infinity at linear convergence rate.
To promise this point, the term $(\gamma)^K \left(C_{\beta}\right)^2$ has to be within $(0, 1)$, which further requires that $\gamma \in (0, 1)$ and $K > - 2 \ln C_{\beta}/ \ln \gamma$. To ensure $\gamma$ to be within $(0,1)$, a proper $\eta$ can be selected;
% \item 
\textbf{2)} The fact that our proposed aggregation method can achieve zero optimality gap is an improvement over the results in {\cite{turan2022robust, wu2020federated, pillutla2022robust}}, which do not prove the zero optimality gap for their proposed aggregation method.
%\end{itemize}
\end{rem}

Based on Theorem \ref{theo:fix}, we can further extend the discussion to general case, which is given as follows.

\begin{theorem} \label{theo:vari}
With a general $\eta_m^{t,k}$ and $K^t$ for any honest user $m$, such that $m \in \mathcal{M} \setminus \mathcal{B}$ at the $k$th update of $t$th round of iteration,
% with variable learning rate $\eta_m^{t,k}$ (\textcolor{red}{or $\eta_m^{t,k}$}) and aggregation step $K^t$,
there is
\begin{small}
 % \begin{align}
 \begin{equation}
    \mathbb{E}  \{ F(\bm{w}^{t + 1}) - F(\bm{w}^*) \} 
    \leqslant \frac{L}{2} \left\lVert \bm{w}^1 - \bm{w}^* \right\rVert^2 \prod_{i=1}^t \sum_{m \in \mathcal{M} \setminus \mathcal{B}} \frac{(C_{\beta})^2}{M - B} \prod_{k=1}^{K^i} \gamma_m^{t, k}
% \end{align}
\end{equation}
  \end{small}
  where $\gamma_m^{t, k} = 1 - 2\eta_m^{t,k} \mu + (\eta_m^{t,k})^2 L^2 (1+\delta^2)$.
\end{theorem}

\begin{IEEEproof}
% Please refer to Appendix \ref{app:vari}.
The proof of Theorem \ref{theo:vari} relies on the following lemma.

\begin{lemma} \label{lem:varib}
%For any honest user $m$, i.e., $\forall m \in \mathcal{M} \setminus \mathcal{B}$ with $\eta_m^{t,k}$ and  aggregation step $K^t$ of time slot $t$,
With a general $\eta_m^{t,k}$ and $K^t$ for any honest user $m$, such that $m \in \mathcal{M} \setminus \mathcal{B}$ at $k$th update of $t$th round of iteration,
$\mathbb{E} \{ \left\lVert \bm{w}_{m}^{t, K^t} - \bm{w}^* \right\rVert^2 \}$ is bounded as
\begin{small}
    \begin{equation}
      \mathbb{E} \left \{ \left\lVert \bm{w}_m^{t, K^{t}} - \bm{w}^* \right\rVert^2 \right \} \leqslant \mathbb{E} \left \{ \left\lVert \bm{w}^t - \bm{w}^* \right\rVert^2 \right \} \prod_{k=1}^{K^i} \gamma_m^{t, k}
    \end{equation}
    \end{small}
  \end{lemma}

  \begin{IEEEproof}
  Please refer to the proof of Lemma \ref{lem:bound}.
  \end{IEEEproof}

  Similar to the proof of Theorem \ref{theo:fix}, we have
\begin{small}
  \begin{align} \label{e:iterative_form}
    &\quad \mathbb{E} \left \{ \left\lVert \bm{w}^{t + 1} - \bm{w}^* \right\rVert^2 \right \} 
    \leqslant \frac{(C_{\beta})^2}{M-B} \mathbb{E} \left \{ \left\lVert \bm{w}^t - \bm{w}^* \right\rVert^2 \right \} \sum_{m \in \mathcal{M} \setminus \mathcal{B}} \prod_{k=1}^{K^i} \gamma_m^{t, k}
  \end{align}
\end{small}
  Iteratively calculating the formula of (\ref{e:iterative_form}), we have
  \begin{small}
  \begin{subequations}
    \begin{align}
      &\quad \mathbb{E} \left \{ \left\lVert \bm{w}^{t + 1} - \bm{w}^* \right\rVert^2 \right \}\nonumber \\
      &\leqslant \left( \frac{(C_{\beta})^2}{M-B} \right)^t \mathbb{E} \left \{ \left\lVert \bm{w}^1 - \bm{w}^* \right\rVert^2 \right \} \prod_{i=1}^t \sum_{m \in \mathcal{M} \setminus \mathcal{B}} \prod_{k=1}^{K^i} \gamma_m^{t, k} \\
      &= \left\lVert \bm{w}^1 - \bm{w}^* \right\rVert^2 \prod_{i=1}^t \sum_{m \in \mathcal{M} \setminus \mathcal{B}} \frac{(C_{\beta})^2}{M - B} \prod_{k=1}^{K^i} \gamma_m^{i, k} \label{inequ:gammai}
    \end{align}
  \end{subequations}
  \end{small}
Then there is
% optimality gap $\mathbb{E} \{ F(\bm{w}^{t + 1}) - F(\bm{w}^*) \}$ has can be shown as
\begin{small}
%\begin{align}
\begin{equation}
    \mathbb{E}  \{ F(\bm{w}^{t + 1}) - F(\bm{w}^*) \} \leqslant \frac{L}{2} \left\lVert \bm{w}^1 - \bm{w}^* \right\rVert^2 \prod_{i=1}^t \sum_{m \in \mathcal{M} \setminus \mathcal{B}} \frac{(C_{\beta})^2}{M - B} \prod_{k=1}^{K^i} \gamma_m^{i, k}
%\end{align}
\end{equation}
\end{small}

%So that Theorem \ref{theo:vari} is proved.
This completes the proof of Theorem \ref{theo:vari}.
\end{IEEEproof}

\begin{rem}
%\begin{itemize}
% \item 
\textbf{1)} Similar with the discussion in Remark \ref{r:mark1}, the optimality gap derived in Theorem \ref{theo:vari} can also converge to zero as $t$ tends to infinity at linear convergence rate, when $\sum_{m \in \mathcal{M} \setminus \mathcal{B}} \prod_{k=1}^{K^i} \gamma_m^{i, k} < {(M-B)}/{(C_{\beta})^2}$ holds;
%\item 
\textbf{2)} Compared with {\cite{turan2022robust, wu2020federated, pillutla2022robust}}, even with a general setup of $\eta_m^{t,k}$ and $K^t$, rather than setting them to be a common $\eta$ and $K$ respectively, we can still achieve zero optimality gap;
%\item 
\textbf{3)} Compared with the setup in Theorem \ref{theo:fix}, $\eta_m^{t,k}$ and $K^t$ could be set differently over $t$, $k$, and $m$, which permits more flexibility.
For example, at the early stage of training, we do not wish $K^t$ to be too high, which may lead the returned $\bm{z}_m^t$ to be overfitted for the local model of user $m$.
As a comparison, at the ending stage of training, the model parameter $\bm{w}^{t+1}$ are nearly stable, and $K^t$ could be set larger so as to speed up convergence.
% \end{itemize}
\end{rem}

\section{Numerical Results}\label{sec:numer}

In this section, numerical results are presented.
Four methods are investigated for comparison:
1) our proposed method with uniform selection of $K^t$ and $\eta_m^{t,k}$, denoted as $K$ and $\eta$ respectively.
2) our proposed method with individual selection of $K^t$ and $\eta_m^{t,k}$;
3) the method in \cite{pillutla2022robust};
4) the method in \cite{turan2022robust} \footnote{{The method in \cite{wu2020federated} has nearly the same performance with \cite{turan2022robust} and hence omitted due to limited space}}.
These four methods are denoted as Proposed-Uniform, Proposed-General, RFA, and RANGE, respectively.
%, for the ease of presentation.
The task is to train an image classifier over dataset EMNIST on PyTorch.
%The total number of training rounds 
$E=4000$ and $M=50$.
%For each method, $M=50$ and the 
The batch size is set as 512.
Each Byzantine attacker draws its message from a Gaussian distribution as given in  \cite{pillutla2022robust}.
For RANGE, $K=1$.
For Proposed-Uniform and RFA, $K=6$.
For the Proposed-General method, $K^t = K^1(1-\lfloor t/E \rfloor)$ and $K^1=8$.
For $\eta$, Proposed-Uniform shares the setup as in RFA \cite{pillutla2022robust}, and Proposed-General selects the one as in RANGE \cite{turan2022robust}.

\begin{figure}[htbp]
  \centering
  \subfloat[$\beta = 0.2$]{
    \begin{minipage}[b]{0.22\textwidth}
      \centering
      \includegraphics[angle=0, width=1 \textwidth]{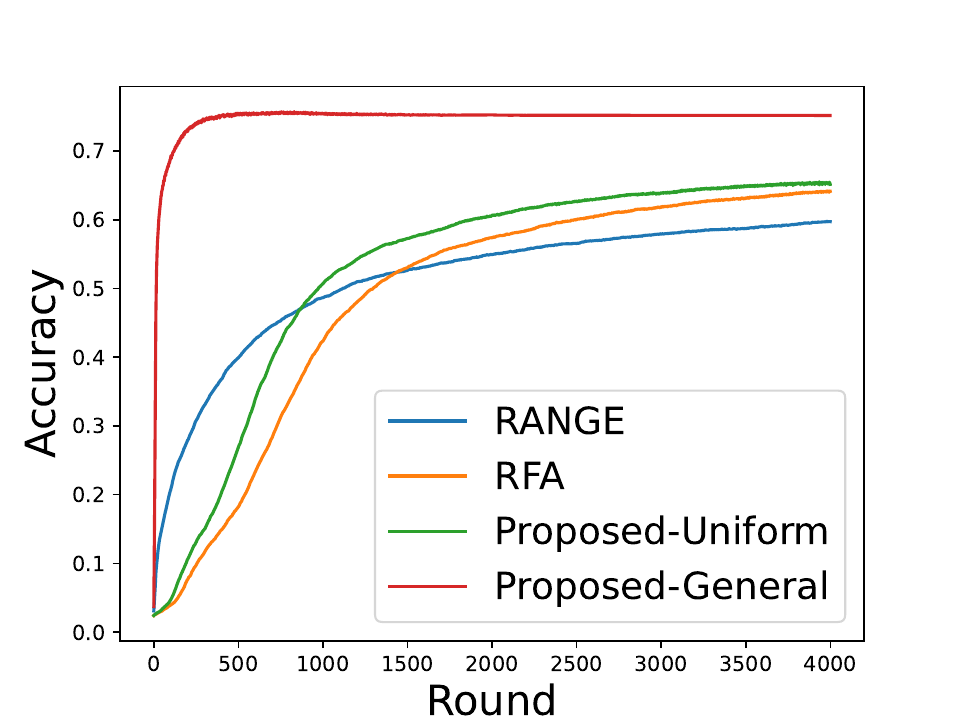}
    \end{minipage}
  }
  \subfloat[$\beta = 0.4$]{
    \begin{minipage}[b]{0.22\textwidth}
      \centering
      \includegraphics[angle=0, width=1 \textwidth]{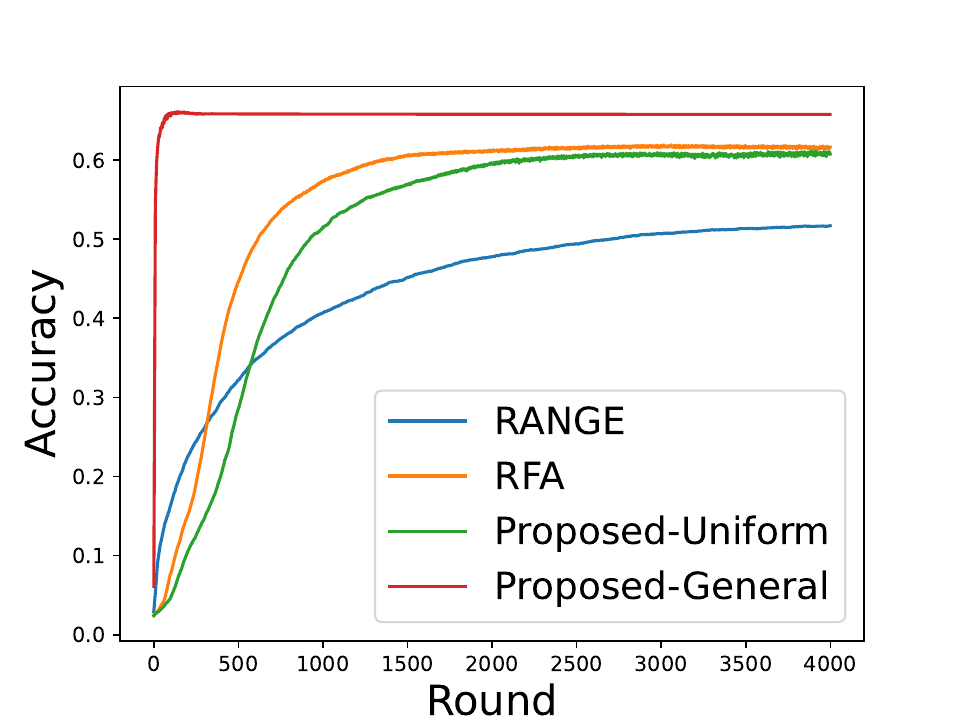}
    \end{minipage}
  }

  \subfloat[$\beta = 0.2$]{
    \begin{minipage}[b]{0.22\textwidth}
      \centering
      \includegraphics[angle=0, width=1 \textwidth]{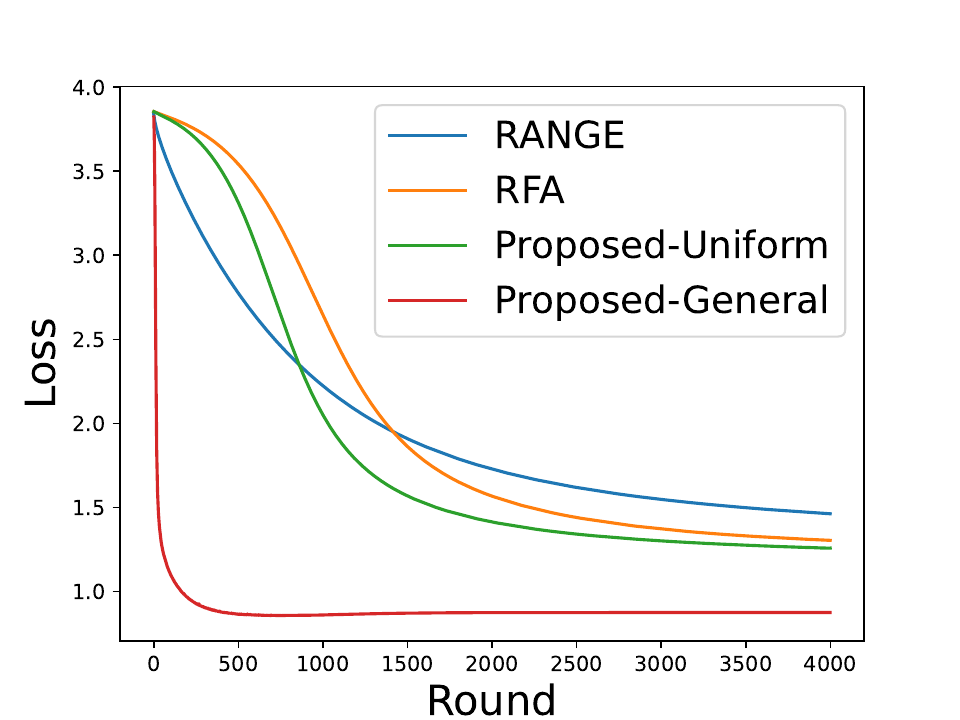}
    \end{minipage}
  }
  \subfloat[$\beta = 0.4$]{
    \begin{minipage}[b]{0.22\textwidth}
      \centering
      \includegraphics[angle=0, width=1 \textwidth]{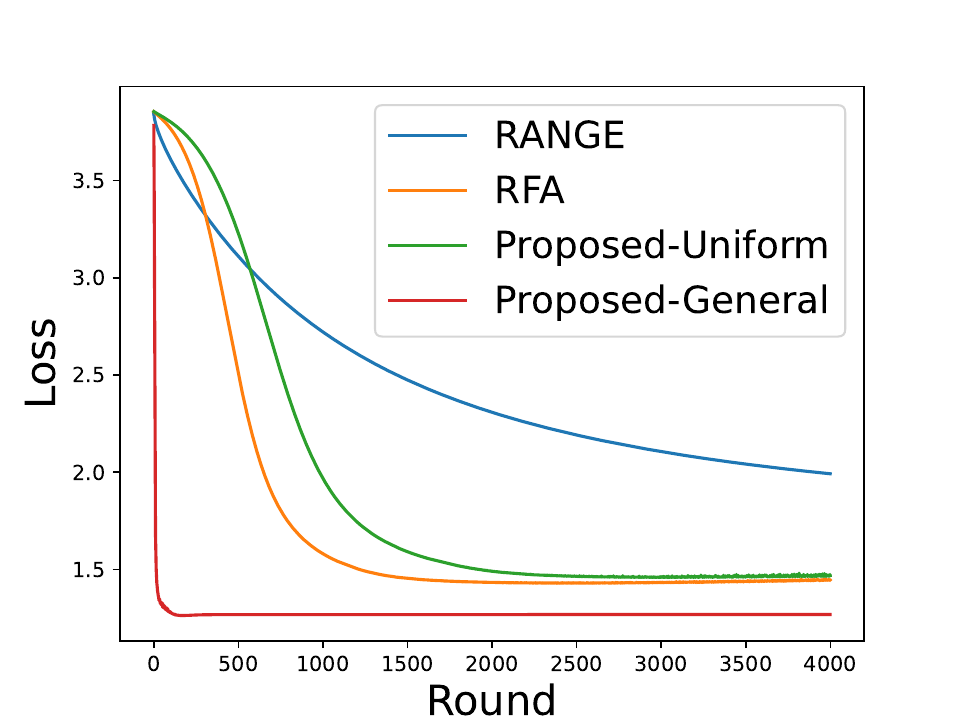}
    \end{minipage}
  }
  \centering
  \caption{Test accuracy and training loss with different $\beta$.}
  \label{pic:per:emnist}
 \end{figure}

\begin{figure}[htbp]
  \centering
\subfloat[Test Accuracy]{
\begin{minipage}[b]{0.45 \linewidth}
\centering
\includegraphics[angle=0, width=1 \textwidth]{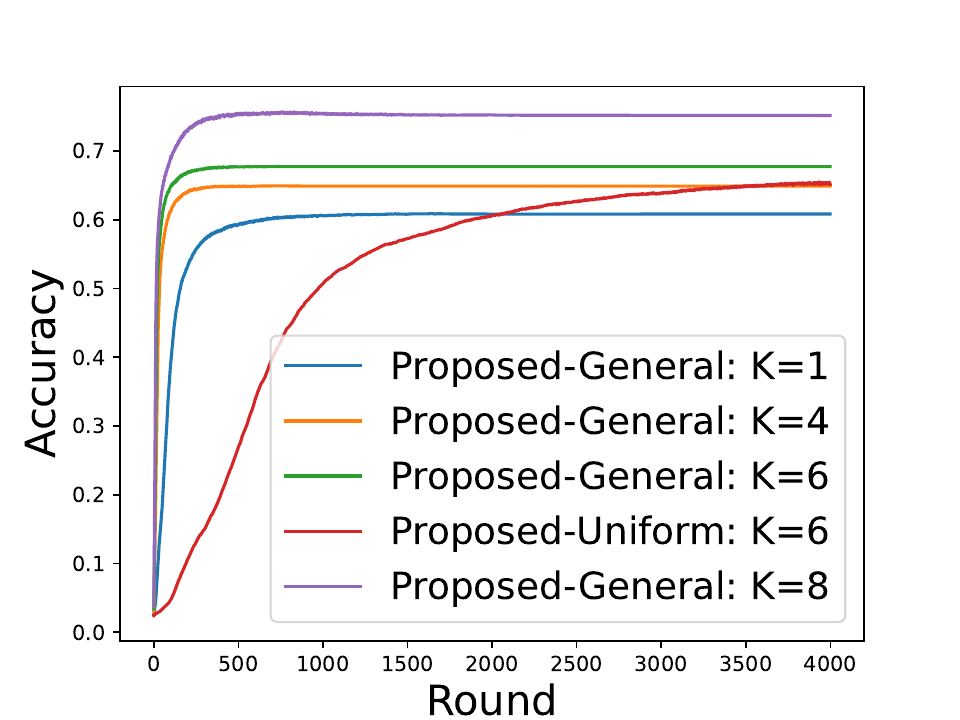} \\
%\centering{(a) Test Accuracy}
    \end{minipage}
    }
\subfloat[Training Loss]{
\begin{minipage}[b]{0.45 \linewidth}
\centering
\includegraphics[angle=0, width=1  \textwidth]{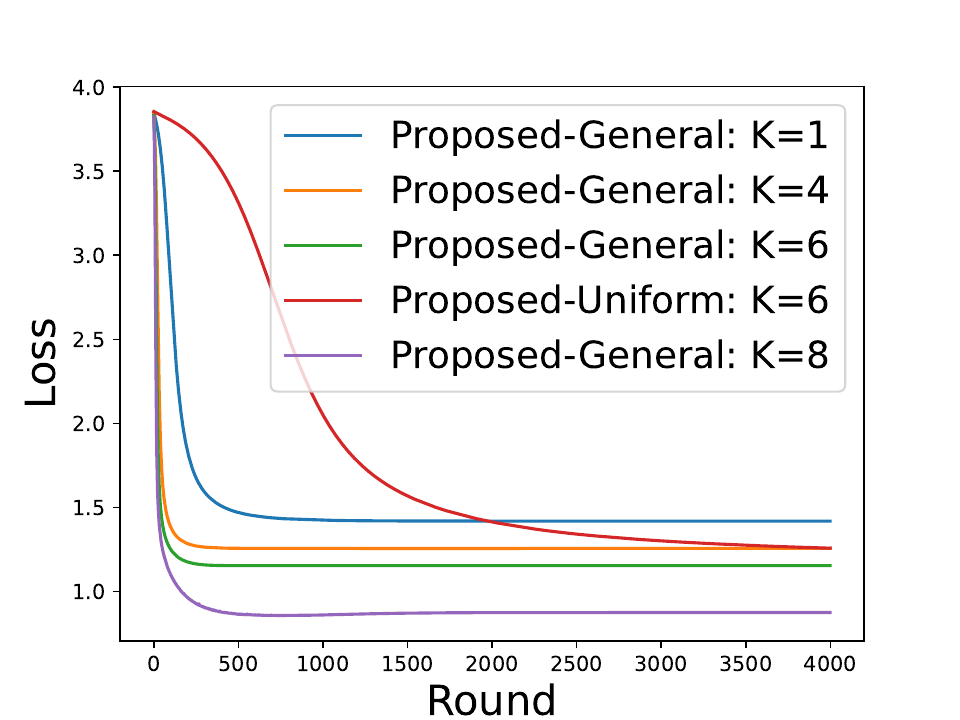} \\
%\centering{(b) Training Loss}
\end{minipage}
}
\centering
\caption{Test accuracy and training loss with different $K$.}  \label{pic:lr:emnist}
\end{figure}

In Fig. \ref{pic:per:emnist}, test accuracy and training loss are plotted respectively for the four compared methods with different selection of $\beta$ value.
%, which corresponds to the ratio of Byzantine attackers. 
As shown in Fig. \ref{pic:lr:emnist}, we plot the test accuracy and the training loss for different values of $K$ with $\beta=0.2$ .
From Fig. \ref{pic:per:emnist} and Fig. \ref{pic:lr:emnist}, we can have the following observations:
1) The Proposed-General method can always achieve less training loss, higher test accuracy and convergence speed. The verifies the effectiveness of our proposed method; 
2) The Proposed-General method can achieve better performance than that of the Proposed-Uniform method, which validates the advantage of flexible selection of $K^t$ and $\eta_m^{t,k}$;
3) Our proposed method can be robust against the Byzantine attack when $\beta$ reaches up to 0.4;
4) Compared with case of $K=1$ as implemented in RANGE method, more steps of local updating could  improve the training performance.

\section{Conclusion}\label{sec:conclu}

In this paper, we propose a communication-and-computation efficient FL aggregation method robust to Byzantine attacks. 
%Our proposed method allows individual setting of learning parameters for every step of local updating, which is shown to be beneficial compared with a uniform setup. 
For our proposed aggregation method, we prove it can achieve zero optimality gap with linear convergence rate as long as the number of Byzantine attackers is less than half of crew users. Our proposed method is also shown to outperform benchmark methods on training performance by numerical results.
%Numerical results further verifies the advantage of our proposed method over benchmark methods on training performance.

\newpage

\bibliography{references}
\bibliographystyle{IEEEtran}
\end{document}